\documentclass[sigconf,natbib=true]{acmart}
\AtBeginDocument{%
  }

\setcopyright{acmlicensed}
\copyrightyear{2018}
\acmYear{2018}
\acmDOI{XXXXXXX.XXXXXXX}
\acmConference[Conference acronym 'XX]{Make sure to enter the correct
  conference title from your rights confirmation email}{June 03--05,
  2018}{Woodstock, NY}
\acmISBN{978-1-4503-XXXX-X/2018/06}

\usepackage{graphicx,hyperref}
\usepackage{tabularx}
\usepackage{booktabs}
\usepackage{makecell}
\usepackage{multirow}
\usepackage{pifont}
\usepackage{xcolor}
\usepackage{placeins}
\usepackage{caption}
\usepackage{float}

\begin{document}

\title{CSyMR: Benchmarking Compositional Music Information Retrieval in Symbolic Music Reasoning}

\author{Boyang Wang}
\affiliation{%
  \institution{University of California, San Diego}
  \city{San Diego}
  \state{California}
  \country{USA}
}

\author{Yash Vishe}
\affiliation{%
  \institution{University of California, San Diego}
  \city{San Diego}
  \state{California}
  \country{USA}
}

\author{Xin Xu}
\affiliation{%
  \institution{University of California, San Diego}
  \city{San Diego}
  \state{California}
  \country{USA}
}

\author{Zachary Novack}
\affiliation{%
  \institution{University of California, San Diego}
  \city{San Diego}
  \state{California}
  \country{USA}
}

\author{Xunyi Jiang}
\affiliation{%
  \institution{University of California, San Diego}
  \city{San Diego}
  \state{California}
  \country{USA}
}

\author{Julian McAuley}
\affiliation{%
  \institution{University of California, San Diego}
  \city{San Diego}
  \state{California}
  \country{USA}
}

\author{Junda Wu}
\affiliation{%
  \institution{University of California, San Diego}
  \city{San Diego}
  \state{California}
  \country{USA}
}

\renewcommand{\shortauthors}{Wang et al.}


\begin{abstract}
Natural language information needs over symbolic music scores rarely reduce to a single step lookup.
Many queries require compositional Music Information Retrieval (MIR) that extracts multiple pieces of evidence from structured notation and aggregates them to answer the question.
This setting remains challenging for Large Language Models due to the mismatch between natural language intents and symbolic representations, as well as the difficulty of reliably handling long structured contexts.
Existing benchmarks only partially capture these retrieval demands, often emphasizing isolated theoretical knowledge or simplified settings.
We introduce \textbf{CSyMR-Bench}, a benchmark for compositional MIR in symbolic music reasoning grounded in authentic user scenarios.
It contains $126$ multiple choice questions curated from community discussions and professional examinations, where each item requires chaining multiple atomic analyses over a score to derive implicit musical evidence.
To support diagnosis, we provide a taxonomy with six query intent categories and six analytical dimension tags.
We further propose a tool-augmented retrieval and reasoning framework that integrates a ReAct-style controller with deterministic symbolic analysis operators built with \textit{music21}.
Experiments across prompting baselines and agent variants show that tool-grounded compositional retrieval consistently outperforms Large Language Model-only approaches, yielding $5$--$7\%$ absolute accuracy gains, with the largest improvements on analysis-heavy categories.
\end{abstract}



\received{20 February 2007}
\received[revised]{12 March 2009}
\received[accepted]{5 June 2009}


\maketitle

\section{Introduction}


Large Language Models (LLMs) have recently shown strong capabilities in music-related tasks, including generation and question answering \cite{yuan-etal-2024-chatmusician,yu2023musicagentaiagentmusic}. For symbolic music, LLMs can operate on text-compatible encodings and achieve competitive performance on music theory style problems with prompting strategies such as in-context learning and chain of thought reasoning \cite{yuan-etal-2024-chatmusician,pond2025teachingllmsmusictheory}. However, many real user information needs over scores are not well modeled by single-step recognition or knowledge recall. They instead resemble a structured retrieval problem where the answer is implicit in the score and must be derived by composing multiple evidence extraction steps from symbolic notation. This compositional setting is challenging because evidence is distributed across heterogeneous musical dimensions, and symbolic contexts can be long and highly structured, making direct end-to-end reasoning challenging and limited to ungrounded intermediate conclusions.

\begin{figure}[ht]
  \centering
  \includegraphics[width=0.95\linewidth]{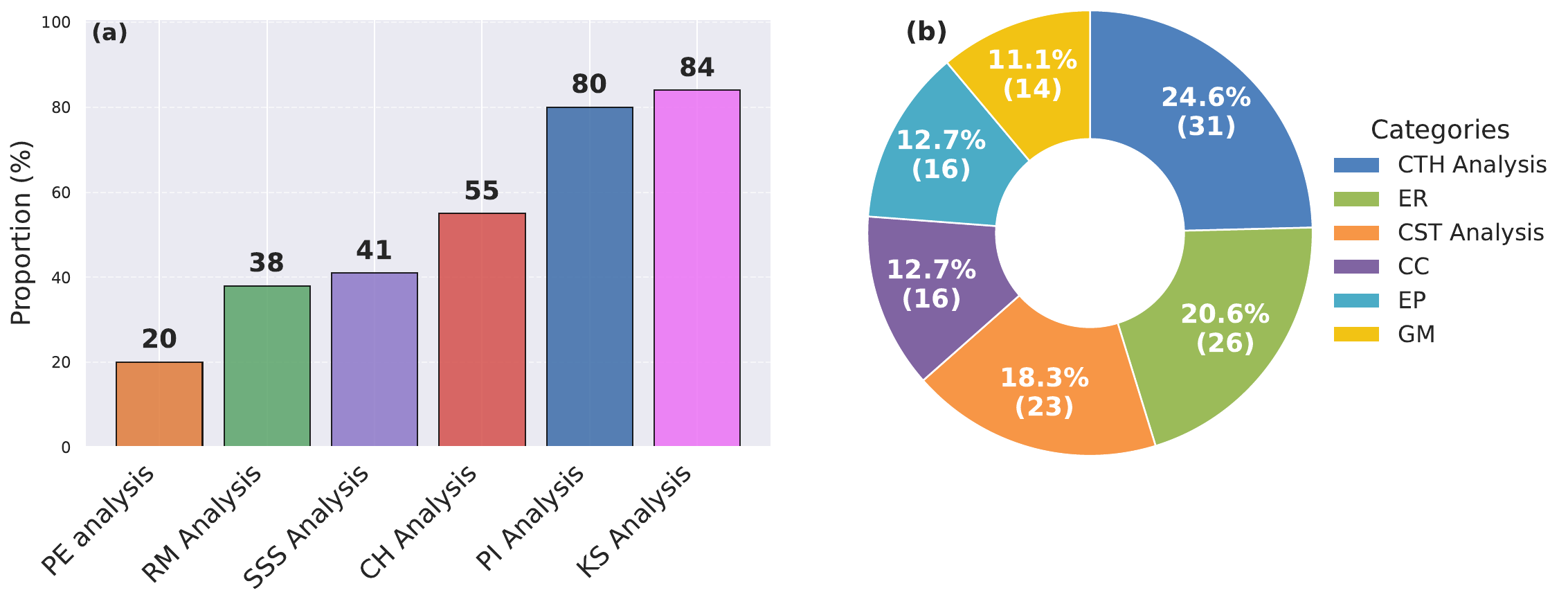}
  \vspace{-0.8em}
  \caption{Statistics of CSyMR-Bench. (a) shows the frequency of each Analytical Dimension. Abbreviations: PE (Performance \& Expression), RM (Rhythm \& Meter), SSS (Score Structural Statistics), CH (Chord \& Harmony), PI (Pitch \& Interval), and KS (Key \& Scale). (b) shows the distribution of Query Intent Categories. Abbreviations: CTH Analysis (Complex Tonal-Harmonic Analysis), ER (Editing / Rewriting), CST Analysis (Complex Structural-Textural Analysis), CC (Composition \& Creative Guidance), EP (Effect / Perceptual Explanation), and GM (Genre / Musician Judgment).}
  \label{fig:cat}
  \vspace{-0.8em}
\end{figure}


Existing benchmarks provide valuable foundations but do not fully capture this retrieval-oriented view of symbolic music. Some datasets primarily evaluate music theory knowledge or single-step analysis \cite{yuan-etal-2024-chatmusician,pond2025teachingllmsmusictheory}. Others rely on synthetic symbolic problems that cover fewer authentic query intents and simpler score contexts \cite{wang2025synthesizingsheetmusicproblems}. In-the-wild collections such as WildScore focus on sheet music images \cite{mundada2025wildscorebenchmarkingmllmsinthewild} rather than symbolic formats, while broader suites emphasize audio perception or general multimodal understanding \cite{ma2025cmibenchcomprehensivebenchmarkevaluating,li-etal-2024-music,DBLP:conf/ismir/WeckMBQFB24,ma2025mmarchallengingbenchmarkdeep}. As a result, a gap remains for benchmarks that target compositional retrieval and evidence aggregation over symbolic scores in realistic user scenarios.

\begin{figure*}[t]
  \centering
  \vspace{-1em}
  \includegraphics[width=1\linewidth]{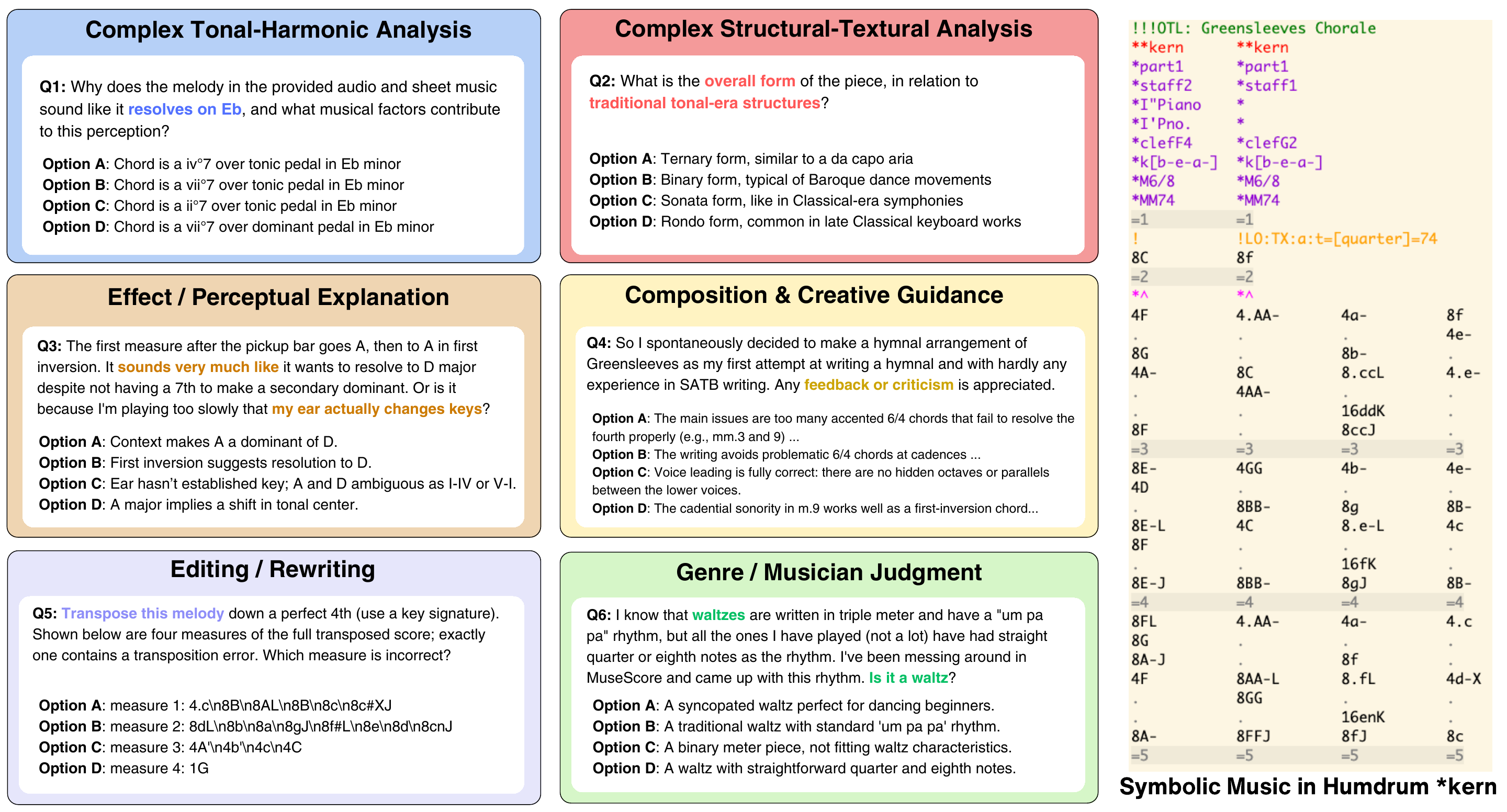}
  \caption{Representative examples of benchmark questions across different categories. One question is shown for each category; in addition, an excerpt of the corresponding symbolic music for Question 4 is displayed in Humdrum kern format.}
  \label{fig:example}
  \vspace{-0.6em}
\end{figure*}


To bridge this gap, we introduce \textbf{CSyMR-Bench}, a benchmark for compositional Music Information Retrieval in symbolic music reasoning. CSyMR-Bench contains $126$ curated multiple-choice questions drawn from community discussions and professional examinations. Each question requires chaining multiple atomic analyses over a symbolic score to recover implicit evidence. To support diagnosis beyond aggregate accuracy, we annotate each item with a taxonomy of six query intent categories and six analytical dimension tags, summarized in Figure~\ref{fig:cat}, with representative examples in Figure~\ref{fig:example}.

We further present a tool-augmented retrieval and reasoning framework that integrates a ReAct-style controller with deterministic symbolic analysis operators built with \textit{music21} \cite{DBLP:conf/ismir/CuthbertA10}. The framework decomposes a query into atomic operations on the score, executes them through verifiable tools, and uses the returned evidence to complete the reasoning chain. This design treats symbolic analysis functions as structured retrieval operators, improving reliability on multi-step questions while keeping the overall task interface unchanged.
We summarize our contributions as follows: \textbf{(1)} We release CSyMR-Bench, which captures the complexity of real-world music inquiries. \textbf{(2)} We provide a fine-grained taxonomy for diagnosing reasoning capabilities. \textbf{(3)} We demonstrate that a tool-augmented agent significantly outperforms baselines by grounding LLM reasoning in verifiable information extraction from scores.

\section{The CSyMR Benchmark}
\subsection{Task Definition}
We formulate the task as Compositional MIR over structured documents. Let $\mathcal{D}$ denote a symbolic music score (represented in Humdrum kern \cite{10.5555/275928.275976}) acting as the structured document, and $\mathcal{Q}$ be a natural language query representing a specific user information need. The objective is to identify the correct answer $a$ from a candidate set $\mathcal{A}$ by grounding the decision in $\mathcal{D}$.

Unlike standard text retrieval where answers often exist as explicit spans, Compositional MIR requires extracting implicit evidence scattered across the score's structure. We define an atomic retrieval operation (or evidence extraction) as $op(\mathcal{D}) \rightarrow e$, where $e$ is a specific musical feature (e.g., a chord root or interval) deterministically inferred from the raw notation. A compositional task necessitates constructing an evidence aggregation path $\mathcal{R} = \{e_1, e_2, \dots, e_n\}$, which synthesizes multiple atomic evidence pieces to bridge the semantic gap between the symbolic document $\mathcal{D}$ and the high-level intent in $\mathcal{Q}$.

\subsection{Data Sources and Curation}

The benchmark comprises two subsets representing complementary information needs:

\paragraph{\textbf{Community-Derived Inquiries.}} To capture genuine information needs in the wild, we curated discussions from \textit{r/musictheory} (2012--2022). Unlike synthetic datasets, these queries reflect real-world user intents where individuals seek explanations for complex musical phenomena. 
Raw score images were filtered using a YOLO-based detector \cite{khanam2024yolov11overviewkeyarchitectural} to ensure document quality, followed by manual selection to retain only tasks requiring multi-step evidence retrieval. Selected images were converted to symbolic Humdrum kern format via optical music recognition (OMR) software \cite{soundslice2025}. Ground truth answers were derived from expert-verified, high-engagement comments, while plausible distractor options were synthesized using GPT-4o-mini \cite{openai2024gpt4omini}. This follows the same pipeline as \cite{mundada2025wildscorebenchmarkingmllmsinthewild}.

\paragraph{\textbf{Expert-Domain Problems.}} To benchmark professional-level symbolic reasoning, we sourced questions from college-level music theory examinations. These items underwent the same OMR conversion pipeline. Open-ended questions were adapted into a multiple-choice format to standardize the retrieval target. Specifically, we employed GPT-4o-mini to summarize the reference answer into the correct option and synthesize plausible distractors. All data sources are explicitly documented.

\subsection{Taxonomy}

Compositional music retrieval typically goes beyond analyzing a single type of musical element. 
Thus, we organize CSyMR-Bench according to practical query intents rather than isolated theoretical concepts. We define six Query Intent Categories (shown in Figure~\ref{fig:example}), categorizing the high-level goal of the user's information need.

In addition, to more precisely capture the analytical dimensions involved in each question, we assign tags denoting the necessary analytical dimensions. These cover six axes: Pitch \& Interval, Chord \& Harmony, Key \& Scale, Score Structural Statistics, Rhythm \& Meter, and Performance \& Expression. A compositional task typically involves cross-dimensional retrieval, requiring the system to retrieve and synthesize evidence across orthogonal musical axes. Figure~\ref{fig:cat} summarizes the distribution of query intent categories and analytical dimensions.

\begin{table*}[t]
\small
\centering
\caption{Per-category accuracy (\%) on CSyMR-Bench across methods. For each category, only averaged accuracy is shown. The rightmost group reports Exam / Reddit / Averaged results (averaged values are weighted across subsets). In this experiment, we use GPT-4.1-mini in all methods. FS = Few-Shot. Bold indicates the best performance and underline indicates the second-best performance for each column (ties allowed).}
\vspace{-0.6em}
\label{tab:result}
\resizebox{\linewidth}{!}{%
\begin{tabular}{lcccccc|ccc}
\toprule
\textbf{Method} &
\makecell{Complex Tonal-\\Harmonic Analysis} &
\makecell{Editing /\\Rewriting} &
\makecell{Effect / Perceptual\\Explanation} &
\makecell{Composition \&\\Creative Guidance} &
\makecell{Complex Structural-\\Textural Analysis} &
\makecell{Genre / Musician\\Judgment} &
\multicolumn{3}{c}{\makecell{Average}} \\
\cmidrule(lr){8-10}
 &  &  &  &  &  &  & \textbf{Exam} & \textbf{Reddit} & \textbf{Avg} \\
\midrule
GPT
& 54.84
& 46.15
& 50.00
& \textbf{68.75}
& 39.13
& \underline{50.00}
& 39.02
& 56.47
& 50.79 \\
GPT-FS
& 54.84
& \underline{65.38}
& 62.50
& \textbf{68.75}
& \textbf{56.52}
& \underline{50.00}
& \textbf{56.10}
& 61.18
& 59.52 \\
CoT
& 61.29
& 57.69
& \textbf{81.25}
& 50.00
& \underline{52.17}
& 42.86
& 48.78
& 62.35
& 57.94 \\
CoT-FS
& 61.29
& \textbf{73.08}
& 68.75
& \underline{62.50}
& \textbf{56.52}
& 42.86
& 48.78
& \underline{68.24}
& \underline{61.90} \\
ReAct
& \underline{67.74}
& 50.00
& \underline{75.00}
& 56.25
& \textbf{56.52}
& \textbf{64.29}
& 48.78
& 67.06
& 61.11 \\
Music21 ReAct
& \textbf{77.42}
& \textbf{73.08}
& \textbf{81.25}
& \textbf{68.75}
& \underline{52.17}
& 35.71
& \underline{53.66}
& \textbf{72.94}
& \textbf{66.67} \\
\bottomrule
\end{tabular}%
}
\end{table*}

\section{Tool-Augmented Retrieval Agent}

Direct reasoning over complex symbolic scores ($\mathcal{D}$) is prone to hallucination due to the impedance mismatch between natural language and structured notation. To address this, we propose a retrieval-augmented framework that integrates a ReAct-style \cite{yao2022react} controller with deterministic MIR operators built with \texttt{music21}. Instead of relying on parametric knowledge, our agent explicitly invokes these operators to perform atomic retrieval ($op(\mathcal{D}) \rightarrow e$), thereby transforming the task into a process of verifiable evidence aggregation.

\subsection{Framework Overview}
\begin{figure}[ht]
  \centering
  \includegraphics[width=1.\linewidth]{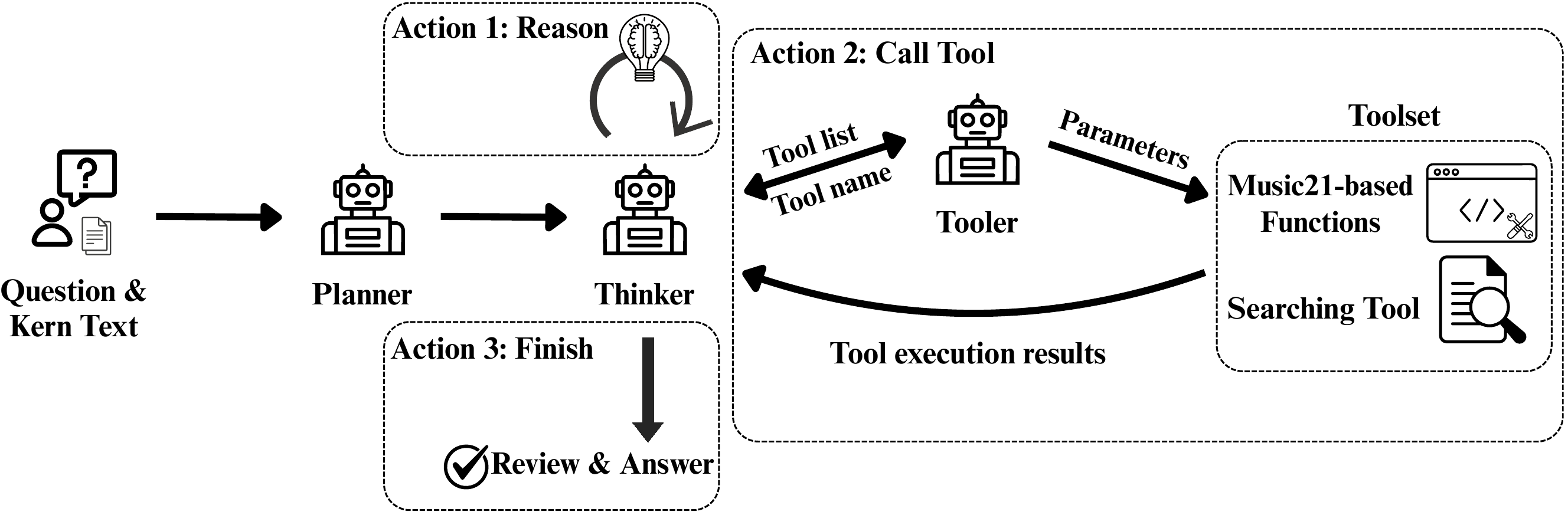}
  \vspace{-1.5em}
  \caption{Overview of the Tool-Augmented Retrieval Agent. The framework iteratively extracts symbolic evidence via deterministic tools to construct a verifiable reasoning chain.}
  \label{fig:framework}
\vspace{-1.5em}
\end{figure}

As shown in Figure~\ref{fig:framework}, the agent operates as an iterative evidence retrieval loop. The Planner decomposes the high-level query $\mathcal{Q}$ into actionable steps, while the Thinker maintains the evidence aggregation path $\mathcal{R}$, dynamically formulating retrieval actions based on the current context. This iterative process allows the system to navigate the score structure step-by-step, validating intermediate findings before deriving the final answer.


\subsection{Deterministic Symbolic Operators}

To ground the reasoning in reliable facts, the system is equipped with a Deterministic Toolset comprising 16 \texttt{music21}-based analysis functions. These act as verifiable retrieval operators across the six dimensions defined in Section 2.3. To ensure robustness, all tools are implemented as strictly typed operators. We enforce a context isolation strategy where raw code execution is hidden from the reasoning agent; a dedicated Tooler executes these operators and returns solely natural-language evidence summaries ($e$). This design guarantees that the agent's attention is strictly allocated to processing retrieved musical evidence rather than debugging code execution.


\begin{table}[H]
\centering
\vspace{-0.5em}
\caption{Accuracy (\%) of different models under the Direct LLM setting. Total values are weighted averages over the exam and Reddit subsets. DS is short for DeepSeek.\\}
\vspace{-1.8em}
\label{tab:direct_llm}
\resizebox{\linewidth}{!}{%
\begin{tabular}{lccccc}
\toprule
\textbf{Subset} & \textbf{GPT-4.1-mini} & \textbf{GPT-4.1} & \textbf{DS-V3.1} & \textbf{Gemini-1.5-Pro} & \textbf{Claude-3.7-Sonnet} \\
\midrule
Exam   & 39.02 & 34.15 & 31.71 & 29.27 & 34.15 \\
Reddit & 56.47 & 54.12 & 55.29 & 54.12 & 61.17 \\
\midrule
\textbf{Total} & 50.79 & 47.62 & 47.62 & 46.03 & \textbf{52.38} \\
\bottomrule
\end{tabular}%
}
\vspace{-1.2em}
\end{table}

\begin{figure*}[t]
  \centering
  \includegraphics[width=0.98\linewidth]{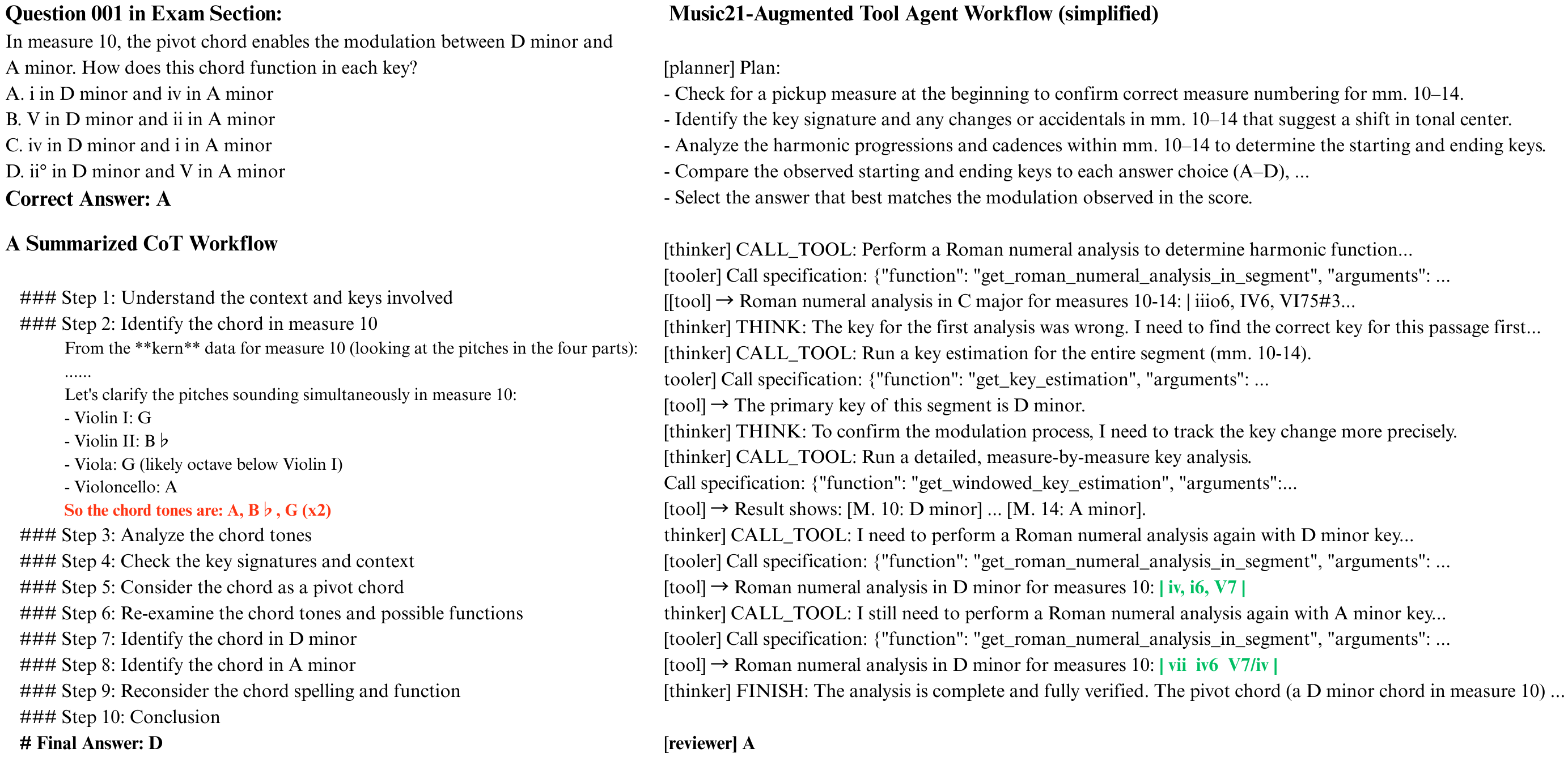}
  \caption{Case study of an example question in the Benchmark. The Music21-Augmented Tool Agent gets the correct answer, while CoT fails.}
  \label{fig:case}
\vspace{-1em}
\end{figure*}

\section{Experiments}
\subsection{Setup}
We evaluate six prompting and reasoning strategies: Standard Zero-shot GPT (directly selecting the answer letter), GPT Few-Shot, Chain-of-Thought (CoT), CoT Few-Shot, Vanilla ReAct without musical tools, and our proposed ReAct-style \texttt{music21} tools-augmented agent framework. All prompting baselines and agent controllers are implemented with \textbf{GPT-4.1-mini} as the backbone model to ensure comparability.

Additionally, to benchmark the intrinsic symbolic processing capabilities of state-of-the-art models, we further evaluate five LLMs under the Standard Zero-shot setting: \textbf{GPT-4.1-mini}, \textbf{GPT-4.1}, \textbf{DeepSeek-V3.1}, \textbf{Gemini 1.5 Pro}, and \textbf{Claude 3.7 Sonnet}.

All models receive identical inputs: symbolic score excerpts ($\mathcal{D}$), a natural-language query ($\mathcal{Q}$), and candidate answers ($\mathcal{A}$). Decoding parameters are consistent across runs (temperature=$0$, max tokens and sampling strategy kept at default value), with the maximum trajectory length for ReAct-based agents capped at 12 steps.


\subsection{Main Results}
Table~\ref{tab:result} summarizes performance across different paradigms. While Few-shot prompting and CoT improve over direct selection, vanilla ReAct fails to surpass CoT, indicating that reasoning structure alone is insufficient without effective grounding. In contrast, our Tool-Augmented Retrieval approach achieves the best overall accuracy, consistently outperforming all parametric baselines.

At the category level, tool augmentation yields distinct gains on analysis-heavy tasks, which confirm the efficacy of score-grounded evidence aggregation. Conversely, improvements are negligible for metadata-driven categories (e.g., Genre Judgment). We attribute this to the toolset's design for explicit structural retrieval, which cannot fully substitute for the implicit parametric knowledge required for stylistic classification.

Table~\ref{tab:direct_llm} highlights an interesting trend: \textbf{GPT-4.1-mini} leads in Zero-shot settings. However, with CoT, \textbf{GPT-4.1} reaches 69.41\%, substantially outperforming the mini model. This suggests that while smaller models are efficient for direct surface-level answering, larger models benefit significantly more from structured reasoning when addressing complex compositional MIR problems.

\subsection{Case Study}

Fig.~\ref{fig:case} illustrates how our framework solves a compositional retrieval problem requiring \emph{key modulation detection} and \emph{roman numeral analysis}. The query asks for the harmonic function of a specific chord under a modulated key.
The Chain-of-Thought baseline generates a plausible-looking rationale but fails due to hallucination—fabricating non-existent pitches—leading to a breakdown in the reasoning chain.

By contrast, our agent constructs a verifiable evidence path: it (i) parses input parameters to restrict the retrieval scope, (ii) invokes a windowed key-estimation operator to detect the true modulation, and (iii) applies a roman-numeral analysis operator for rule-consistent labeling.
This tool-grounded approach enables the agent to recover from initial assumptions and derive the correct answer through deterministic evidence, which bridges the semantic gap where purely probabilistic generation failed.


\vspace{-0.4em}

\section{Conclusion}
In this work, we introduced CSyMR-Bench, the first benchmark dedicated to Compositional Music Information Retrieval over symbolic scores, alongside a Tool-Augmented Retrieval Agent that integrates deterministic operators into the reasoning loop. Our experiments demonstrate that grounding LLM generation in verifiable symbolic evidence significantly outperforms standard reasoning baselines, particularly in rigorous analytical tasks. By effectively bridging the semantic gap between natural language and structured scores, this framework establishes a robust paradigm for trustworthy information seeking in musical domains.

\bibliographystyle{ACM-Reference-Format}
\bibliography{refs}

\appendix

\end{document}